\documentclass[letterpaper]{article} 
\usepackage{aaai2026}  
\usepackage{times}  
\usepackage{helvet}  
\usepackage{courier}  
\usepackage[hyphens]{url}  
\usepackage{graphicx} 
\usepackage{natbib}  
\usepackage{caption} 
\usepackage{algorithm}
\usepackage{algorithmic}

\usepackage{amsmath}
\usepackage{booktabs}
\usepackage{multirow}
\usepackage{tabularx}

\usepackage{newfloat}
\usepackage{listings}
\DeclareCaptionStyle{ruled}{labelfont=normalfont,labelsep=colon,strut=off} 
\floatstyle{ruled}
\newfloat{listing}{tb}{lst}{}
\floatname{listing}{Listing}
\title{FilmSceneDesigner: Chaining Set Design for Procedural Film Scene Generation}
\author {
    Zhifeng Xie\textsuperscript{\rm 1,\rm 2},
    Keyi Zhang\textsuperscript{\rm 1},
    Yiye Yan\textsuperscript{\rm 1},
    Yuling Guo\textsuperscript{\rm 1},
    Fan Yang\textsuperscript{\rm 1},
    Jiting Zhou\textsuperscript{\rm 1,\rm 2},
    Mengtian Li
\textsuperscript{\rm 1,\rm 2}\thanks{Corresponding author.}
}
\affiliations {
    \textsuperscript{\rm 1}Department of Film and Television Engineering, Shanghai University\\
    \textsuperscript{\rm 2}Shanghai Engineering Research Center of Motion Picture Special Effects\\
        \{zhifeng\_xie, ky\_zhang, 2543912909, 1304422167, yangphan, zjting, mtli\}@shu.edu.cn
}

\usepackage{bibentry}

\begin{document}

\maketitle

\begin{abstract}
Film set design plays a pivotal role in cinematic storytelling and shaping the visual atmosphere. However, the traditional process depends on expert-driven manual modeling, which is labor-intensive and time-consuming. To address this issue, we introduce \textbf{FilmSceneDesigner}, an automated scene generation system that emulates professional film set design workflow. Given a natural language description, including scene type, historical period, and style, we design an agent-based chaining framework to generate structured parameters aligned with film set design workflow, guided by prompt strategies that ensure parameter accuracy and coherence. On the other hand, we propose a procedural generation pipeline which executes a series of dedicated functions with the structured parameters for floorplan and structure generation, material assignment, door and window placement, and object retrieval and layout, ultimately constructing a complete film scene from scratch. Moreover, to enhance cinematic realism and asset diversity, we construct \textbf{SetDepot-Pro}, a curated dataset of 6,862 film-specific 3D assets and 733 materials. Experimental results and human evaluations demonstrate that our system produces structurally sound scenes with strong cinematic fidelity, supporting downstream tasks such as virtual previs, construction drawing and mood board creation.
\end{abstract}


\section{Introduction}

\begin{figure}[t]
  \centering
  \includegraphics[width=\linewidth]{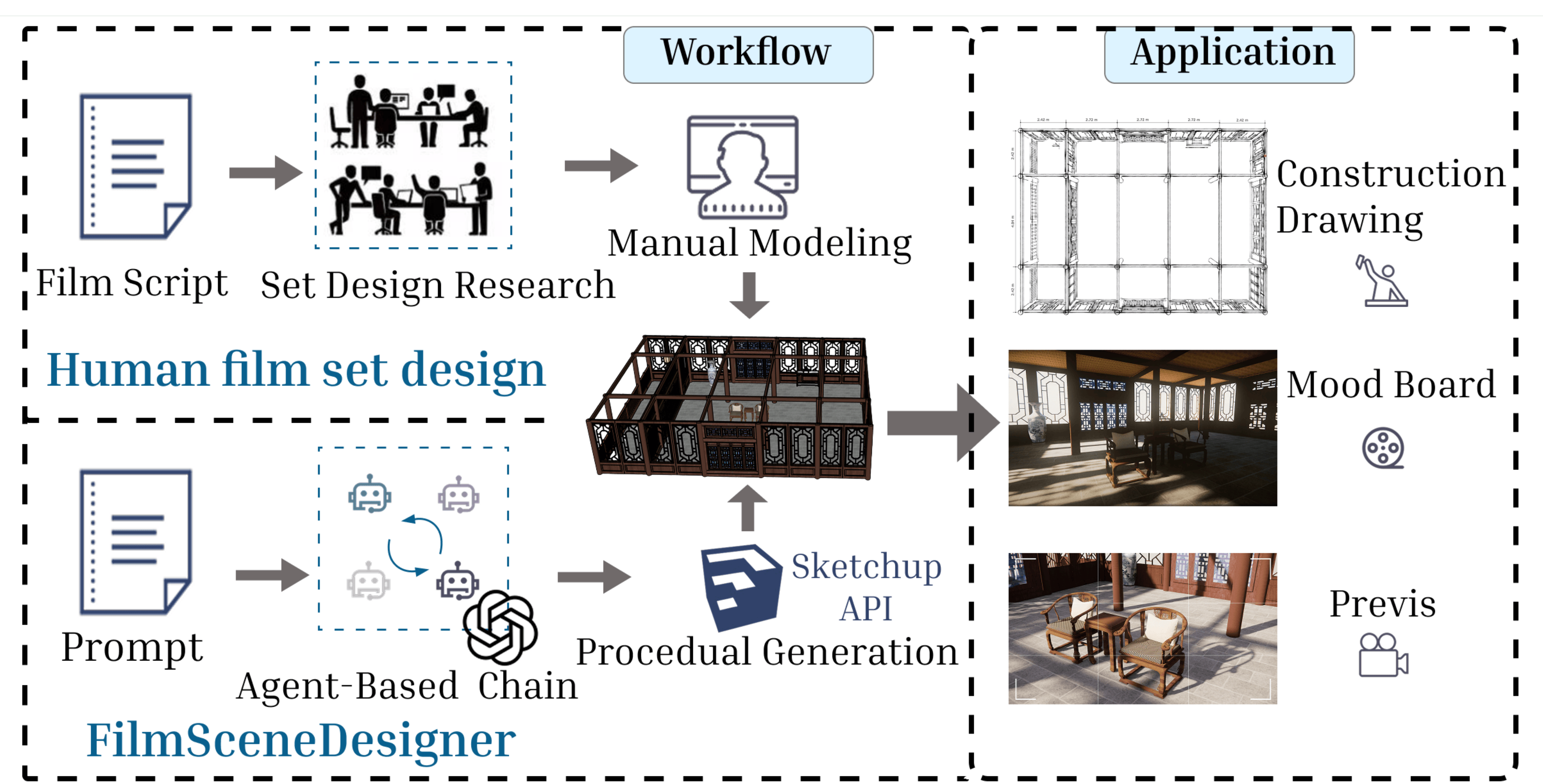}
  \caption{Human Film Set Design: Traditional set design involves labor-intensive script analysis, object research, and manual modeling.
FilmSceneDesigner: Our system leverages an agent chain of set design to automate procedural generation, enabling efficient generation of film sets for construction drawing, mood board, and previs.}
  \label{fig:teaser}
\end{figure}

Film scenes play a crucial role in cinematic storytelling by conveying narrative structure and visual atmosphere. As shown in Figure \ref{fig:teaser}, in traditional workflows, the art department analyzes film scripts, extracting key elements such as scene type, color palette, historical period, and regional characteristics to guide scene creation. After gathering this information, set designers manually construct the scene using professional modeling tools. However, current film production workflows still rely heavily on expert-driven manual modeling, which is time-consuming and requires substantial artistic expertise. This creates a compelling need for an efficient and automated solution tailored to film set design.

Existing scene generation methods fall short in addressing cinematic requirements. Image-based approaches~\cite{chung2023luciddreamer,fang2023ctrl,fridman2023scenescape,hollein2023text2room} rely on depth inference and often produce distorted meshes with artifacts such as holes and stretched surfaces, failing to meet professional standards. Procedural methods~\cite{raistrick2024infinigen} typically require manual parameter configuration and are limited by predefined room templates, lacking structural diversity and flexibility. LLM-based approaches~\cite{yang2024holodeck,ccelen2024design} depend heavily on general-purpose datasets and focus on the functional roles of rooms rather than their structural types (e.g., wall-structure vs. column-structure), resulting in scenes with low cinematic fidelity. In summary, existing approaches suffer from two major challenges when applied to film set design: 
(1) none of them align with the established workflow of film scene design, making it difficult to integrate into production pipelines; 
and (2) they primarily generate general-purpose scenes, lacking the cinematic fidelity required for professional use. 

According to our observations, when professional set designers construct a film scene, the workflow typically involves four sequential stages: (1) constructing the set structure, (2) dressing the surfaces with materials,  (3) installing doors and windows, and (4) arranging props in the scene. In addition, we observe that the quality and appropriateness of material and props play a decisive role in determining whether a scene conveys authentic cinematic realism.

\begin{figure*}[htbp]
  \centering
  \includegraphics[width=1\textwidth]{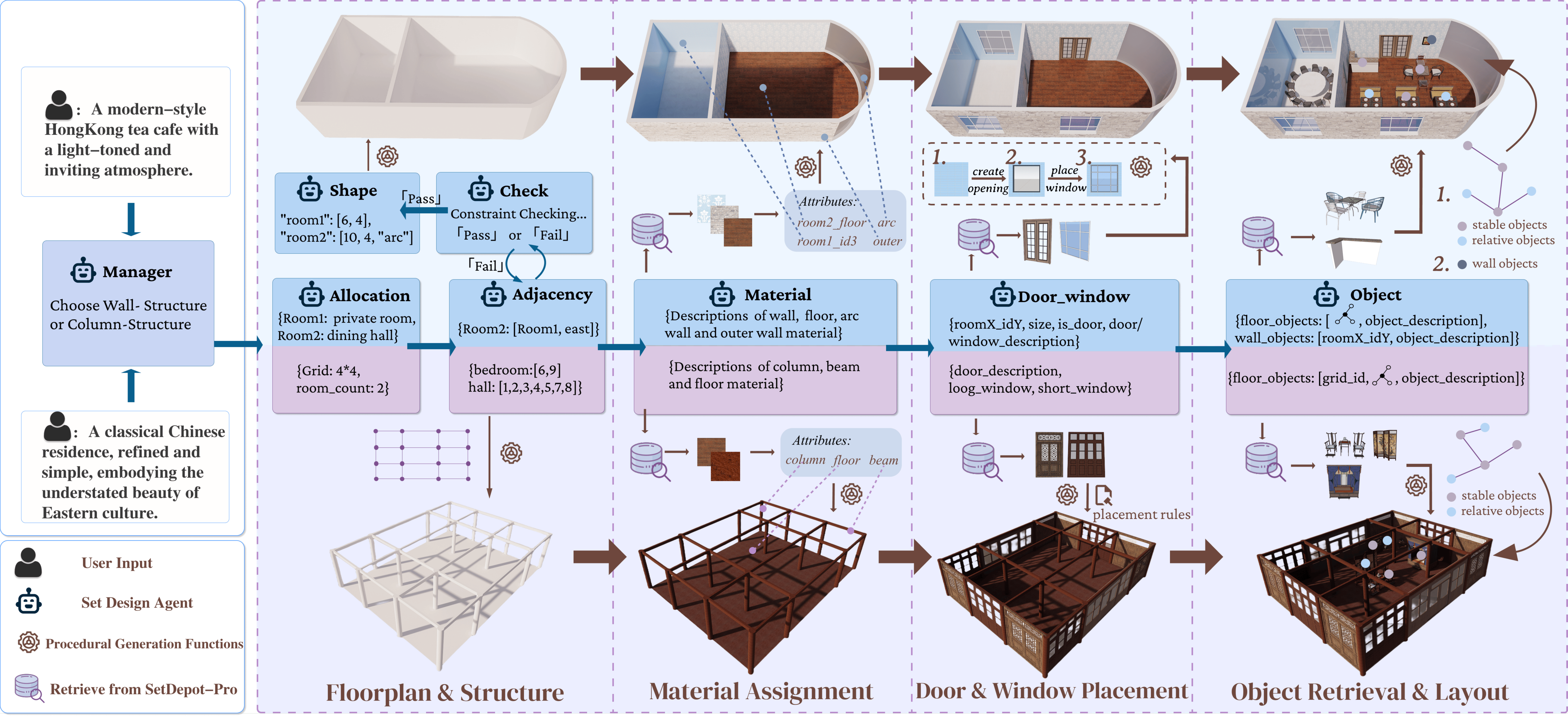}
  \caption{The framework of FilmSceneDesigner. Given a scene description, FilmSceneDesigner constructs an agent-based chaining framework to generate the structured parameters, and then executes the procedural functions with these parameters for floorplan and structure, material assignment, door and window placement, and object retrieval and layout. All required assets, including materials, doors and windows, and objects, are retrieved from SetDepot-Pro to ensure high cinematic fidelity.}
  \label{fig:framework}
\end{figure*}

Inspired by these observations and to address the first challenge, we propose \textbf{FilmSceneDesigner}, a novel system that integrates procedural generation with an agent-based chaining framework to emulate the traditional film set design workflow. As shown in Figure \ref{fig:framework}, given a natural language description including scene type, historical period, and style, the framework first determines the appropriate structural category (wall or column) and generates coherent parameters through prompt-guided reasoning. These parameters are then executed in a procedural pipeline that follows the four key stages of set design: floorplan and structure, material assignment, door and window placement, and object retrieval and layout. Each stage is supported by dedicated generation functions in SketchUp, a widely adopted tool in the film set design industry. Furthermore, our system supports downstream tasks such as virtual previs, construction drawing and mood board creation. To address the second challenge, we construct \textbf{SetDepot-Pro}, a film-specific dataset spanning diverse historical periods and regions to enhance cinematic fidelity.

Our main contributions are as follows:
\begin{itemize}
    \item We propose \textbf{FilmSceneDesigner}, an automated system for film scene generation that emulates professional set design and seamlessly fits production workflows.
    \item We combine procedural generation and an agent-based chaining framework to generate structurally diverse, semantically coherent scenes from natural language.
    \item We construct \textbf{SetDepot-Pro}, a film-specific dataset of 6,862 labeled assets and 733 materials supporting the creation of high-fidelity, stylistically rich film scenes.
\end{itemize}

\section{Related Work}

\subsection{3D Indoor Scene Generation} 
Research on 3D scene generation spans multiple methodologies. \textbf{Image-based} models~\cite{chung2023luciddreamer,fang2023ctrl,fridman2023scenescape,hollein2023text2room} leverage pre-trained text-to-image models and depth estimation to lift 2D images to 3D using NeRF~\cite{mildenhall2021nerf} or Gaussian Splatting~\cite{kerbl20233d} representations. Another major direction is \textbf{layout-driven} scene generation, which defines spatial arrangements before populating a scene. Rule-based methods~\cite{yeh2012synthesizing,weiss2018fast,merrell2011interactive} manually encode spatial constraints, which ensures structural consistency but limits diversity. On the other hand, some methods~\cite{tang2024diffuscene,zhai2023commonscenes,wu2024blockfusion,zhai2024echoscene} learn scene priors from datasets and some of them adapt scene graphs~\cite{gao2024graphdreamer,lin2024instructscene} to represent object relationships and enforce spatial constraints, while they are inherently limited by dataset biases. More recently, \textbf{LLM-driven} approaches have been introduced for scene generation. Layoutgpt~\cite{feng2023layoutgpt} prompts LLMs to directly generate absolute object coordinates and some methods~\cite{yang2024holodeck,fu2024anyhome,ccelen2024design,littlefair2025flairgpt} infer relative spatial relationships and incorporate constraint-based object placement to ensure a plausible layout with LLMs. \textbf{Procedural modeling} is also a powerful tool for scalable 3D scene generation~\cite{raistrick2023infinite}, Infinigen Indoors~\cite{raistrick2024infinigen} employs constraint-based object arrangement to achieve realistic spatial composition. Nevertheless, existing approaches either generate indoor scenes with fixed room types or focus on general-purpose scenes, neither of which meets the requirements of cinematic set design; moreover, they cannot be seamlessly integrated into real film set design workflows.

\subsection{Dataset} 
Datasets for 3D scene generation can be broadly categorized into two types: structured \textbf{indoor scene datasets} and open-vocabulary \textbf{3D asset datasets}. For \textbf{indoor scene}, several datasets provide structured 3D assets specifically for indoor environments. Scan2CAD~\cite{avetisyan2019scan2cad} contains CAD-based synthetic models alongside scanned indoor scenes. 3D-FUTURE~\cite{fu20213dfuture} includes synthetic indoor assets with high-quality textures, enhancing realism in virtual environments. 3D-FRONT~\cite{fu20213dfront} offers synthetic indoor scenes along with their associated digital assets, while ScanNet~\cite{dai2017scannet} consists of scanned indoor scene models. However, these datasets are inherently constrained by predefined room categories, primarily focusing on common household spaces such as bedrooms, kitchens, and living rooms, thereby limiting the diversity of available assets to furniture like tables and chairs. Beyond structured indoor scene datasets, large-scale open-vocabulary \textbf{3D asset datasets} provide a broader range of digital assets. Objaverse~\cite{deitke2023objaverse} aggregates 3D models from diverse sources, offering a vast and heterogeneous collection of objects beyond the constraints of specific room types. Additionally, commercial software platforms such as Unreal Engine Marketplace~\cite{unrealenginemarket}, UE4Arch~\cite{UE4Arch}, and Adobe Stock~\cite{adobestock} maintain their own curated digital asset libraries, supporting various creative and industrial applications. While these datasets provide valuable digital assets, they mainly focus on general-purpose objects. Asset selection plays a crucial role in achieving cinematic realism. To fill this gap, we build a specialized database of 3D assets and materials tailored for film production. Compared to these datasets above, our dataset is specifically designed for film scenes.

\section{Method}

\subsection{Procedural Scene Generation}

\noindent\textbf{Floorplan and Structure.} This stage constructs the foundational floorplan and structural framework of the scene. We formalize two architectural structure types to accommodate diverse structural requirements. We define a wall-structure scene as a collection of rooms \(S\):
\begin{equation}
S = [R_1, R_2, \dots, R_n],
\end{equation} 
where \(R_1, R_2, \dots, R_n\) represent individual rooms within the scene. Each room is composed of a set of boundary edges, which can be either straight lines or arcs, defined as:
\begin{equation}
E_{ij} =
\begin{cases}
[x_{\text{start}}, y_{\text{start}}, x_{\text{end}}, y_{\text{end}}], & \text{line}, \\
[x_{\text{start}}, y_{\text{start}}, x_{\text{end}}, y_{\text{end}}, h_{\text{chord}}], & \text{arc},
\end{cases}
\end{equation}
where \(E_{ij}\) denotes the \(j\)-th edge of the \(i\)-th room. A line is defined by its start and end coordinates, while an arc additionally requires a chord height \(h_{\text{chord}}\). Here, \(x\) and \(y\) are the 2D coordinates.
To represent spatial adjacency relationships between rooms, we use a directed graph representation:
\begin{equation}
A = \{(r_i, r_j, \text{relation}) \mid r_i, r_j \in S, r_i \ne r_j\},
\end{equation}
where \(r_i\) and \(r_j\) denote two rooms, and \textit{relation} specifies their relative positioning, such as east or north. In our modeling system, room layout is based on the 2D coordinate system of SketchUp: the first room (typically \texttt{room1}) is placed at the origin \((0, 0)\), and the positions of subsequent rooms are inferred sequentially based on their spatial relationships with previously placed rooms.

For column-structure scenes, the floorplan is represented as a grid \(G\) of column points:
\begin{equation}
G = [C_{1,1}, C_{1,2}, \dots, C_{m,n}],
\end{equation}
where \(C_{i,j}\) denotes the column located at row \(i\) and column \(j\). Each column is defined by its center coordinates \((x_{i,j}, y_{i,j})\) and radius \(r_{\text{column}}\):
\begin{equation}
C_{i,j} = [x_{i,j}, y_{i,j}, r_{\text{column}}].
\end{equation}

According to the above definitions, we design a series of functions to accomplish the generation process. For wall-structure scenes, we first use the \texttt{parse\_edge} function to infer room boundary edges from room sizes and adjacency relationships. These edges are categorized into two types: external walls and internal walls that partition rooms. We then employ several designed functions to draw straight lines and arcs, followed by the \texttt{add\_face} function to create enclosed floor surfaces from these boundaries. Wall geometry is generated by the \texttt{offset} function to establish inner and outer wall thickness, and finally apply the \texttt{pushpull} function to extrude the walls to the specified height. For column-structure scenes, we compute the center coordinates of all columns based on the grid's row-column count and spacing to form a structured grid. The generation then sequentially applies functions to add the ground plane, instantiate columns at computed positions, and connect adjacent columns with beams to complete the column-beam framework, using the aforementioned functions.

\noindent\textbf{Material Assignment.} This stage assigns appropriate materials to structural components through a systematic attribute-based approach. Our material assignment stage operates in two phases: attribute definition and material application.

In the attribute definition phase, we employ the \texttt{set\_attribute} function to assign unique identifiers to each structural element. For wall-structure scenes represented as $S = \{R_1, R_2, \dots, R_n\}$, we systematically assign attributes to room containers (\texttt{roomX}), floor surface of each room (\texttt{roomX\_floor}), directional inner walls (\texttt{roomX\_idY} where $Y \in \{1,2,3,4\}$ denotes west, south, north, east walls respectively), curved wall segments (\texttt{arc}), and exterior boundaries (\texttt{outer}). For column-structure scenes, attributes are assigned to the ground plane (\texttt{floor}), columns (\texttt{column}), and beams (\texttt{beam}).

\begin{figure}[t]
  \centering
  \includegraphics[width=\linewidth]{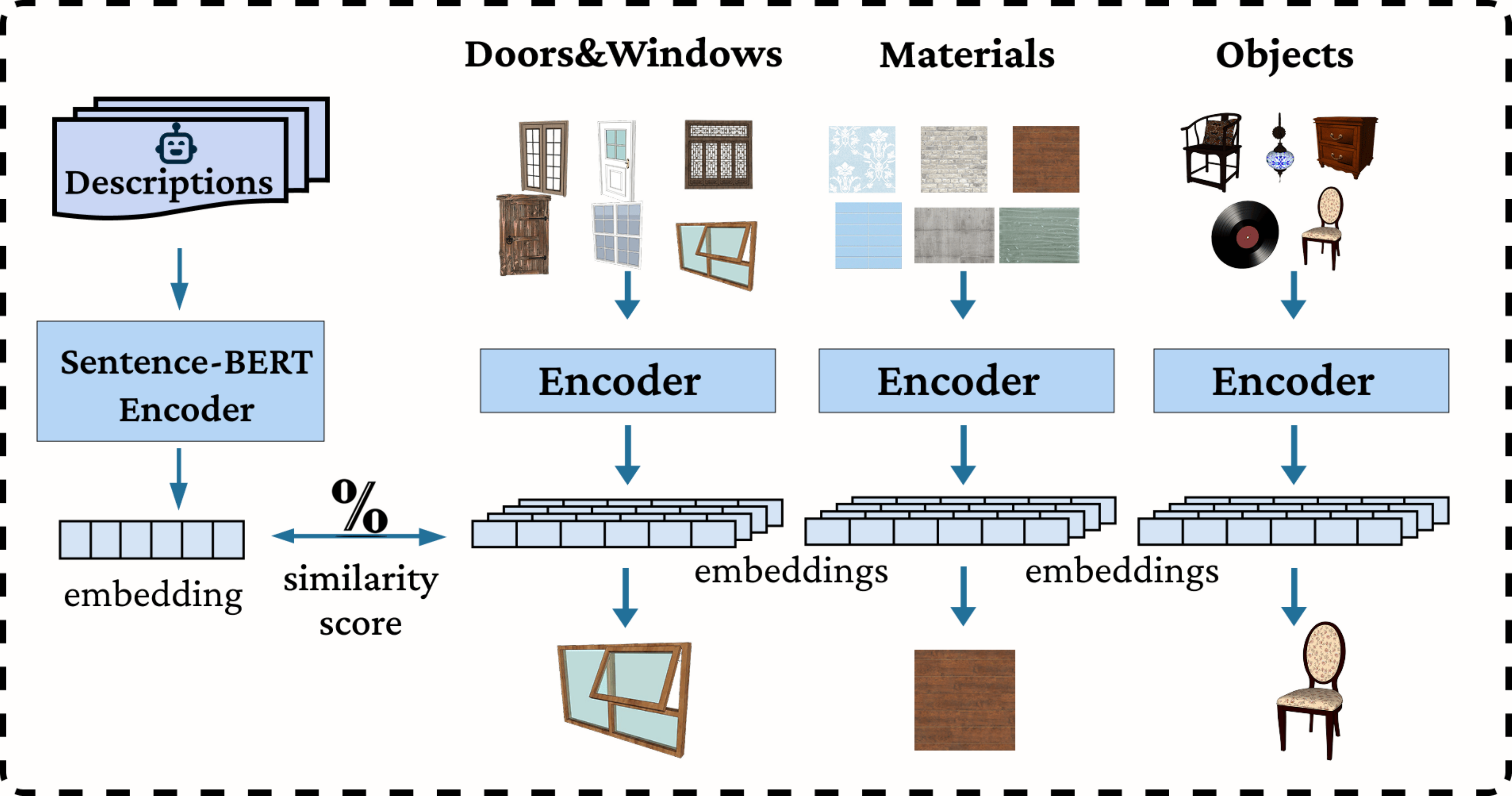}
  \caption{Retrieval Process: Textual descriptions from agent responses are encoded into vector representations using a Sentence-BERT encoder. Similarly, doors, windows, materials, and objects are encoded and stored in an embedding database. By computing similarity scores between the two embeddings, the highest-scoring assets are retrieved.}
  \label{fig:retrieve}
\end{figure}

Following attribute assignment, the \texttt{apply\_material} function performs batch material application by mapping these structured identifiers to corresponding material properties retrieved from our SetDepot-Pro dataset via Sentence-BERT similarity matching. Specifically, as illustrated in Figure~\ref{fig:retrieve}, textual descriptions are encoded into vector representations using a Sentence-BERT~\cite{reimers2019sentence} encoder. Likewise, assets are pre-encoded and stored in an embedding database. Retrieval is performed by computing similarity scores between query and database embeddings, from which the most relevant candidates are selected.

\noindent\textbf{Door and Window Placement.} This stage installs appropriate doors and windows for the current scene. For wall-structure scenes, the process operates in two distinct phases: opening creation and asset placement. In the opening creation phase, we employ the \texttt{open\_wall} function to create openings on target wall surfaces indexed by \texttt{roomX\_idY} attributes. The opening position depends on the asset type: doors are opened from the ground level, while windows are positioned at the wall center.

In the asset placement phase, door and window assets are retrieved from our SetDepot-Pro dataset via Sentence-BERT similarity matching, then positioned through a sequence of dedicated functions: adaptive \texttt{rotate} based on the \texttt{roomX\_idY} wall orientation, \texttt{scale} to match the opening dimensions, and \texttt{translate} to precisely fit the asset into the opening. This two-stage approach ensures accurate geometric alignment between openings and their corresponding assets.

For column-structure scenes, doors and windows are directly inserted into the natural gaps between adjacent columns without requiring opening creation. We categorize placements into three types: doors, long windows, and short windows. Based on extensive analysis of real-world column-structure scenes, we establish a set of placement rules to guide door and window placement. (1) doors are positioned between the middle column pairs of the first and last rows; (2) long windows are placed between the middle column pairs of the first and last columns; (3) remaining perimeter gaps are filled with short windows; (4) short windows are used to partition the entire scene into different rooms. All assets are retrieved from our dataset via similarity matching and positioned with appropriate functions, including rotation, scaling, and translation consistent with the wall-structure scene.

\noindent\textbf{Object Retrieval and Layout.}  
This stage is responsible for selecting semantically appropriate assets and arranging them into spatially coherent layouts. Objects are first retrieved from our SetDepot-Pro dataset using semantic similarity matching. Then the retrieved objects are subsequently arranged according to the underlying structural type. In wall-structure scenes, each room (\texttt{roomX}) serves as the placement unit for both floor and wall objects, while in column-structure scenes, units \(U\) are defined as rectangular regions enclosed by four columns:
\begin{equation}
U = \{ C_{i_1,j_1},\ C_{i_1,j_2},\ C_{i_2,j_1},\ C_{i_2,j_2} \}.
\end{equation}
Only floor objects are placed within each column-structure unit. Floor objects are further categorized into \textit{stable objects} and \textit{relative objects}. Stable objects include corner, edge, and center placements, with edge and center serving as \textit{anchor objects}. Each anchor is assigned a unique \texttt{anchor\_ID}, which can be referenced by relative objects. The placement of a relative object is determined by its anchor reference, combined with a spatial relation (e.g., \textit{left}, \textit{right}, \textit{in front of}, \textit{behind}, \textit{above}) and a distance level (e.g., \textit{near}, \textit{far}). Formally, layout relations are represented as a triplet \(\mathcal{L}\):
\begin{equation}
\mathcal{L} = (o_a, r, o_r),
\end{equation}
where $o_a \in \{\textit{edge}, \textit{center}\}$ is the anchor object, $o_r$ the relative object, and $r=(s,d)$ the composite relation consisting of spatial relation $s$ and distance level $d$. The relative position \(p\) is inferred as:
\begin{equation}
p(o_r) = p(o_a) + \lambda(d)\cdot \vec{v}(s),
\end{equation}
where $\vec{v}(s)$ denotes the direction vector of $s$, and $\lambda(d)$ is the scaling factor associated with $d$. Based on this inference, the system applies the previously introduced placement functions to perform model importation, rotation, scaling, and translation. After these operations, two refinement functions, \texttt{avoid\_collision} and \texttt{refine\_orientation}, are invoked to further adjust object placement, ensuring collision-free layouts. 

For wall-structure scenes, wall objects follow a dedicated strategy: each object is assigned to a wall surface identified by \texttt{roomX\_idY}, adaptively rotated to face the room interior, then scaled and translated to the center of target wall, resulting in spatial plausibility and visual uniformity.

\subsection{Agent-Based Chaining Framework}

\noindent\textbf{Set Design Agents.} To support procedural generation at the conceptual design stage, we construct an agent-based chaining framework that progressively transforms natural language descriptions into structured parameters required by procedural generation functions. Within this framework, multiple set design agents are defined, each responsible for a specific sub-task. Following the workflow decomposition in professional film set design, the complex generation task is divided into sub-tasks with clearly defined responsibilities, each handled by one or more dedicated agents. The overall process is initiated and coordinated by the \textit{Manager}, which sequentially invokes the agents to complete the four stages of procedural generation. The duties of all set design agents are summarized in Table~\ref{tab:agent_roles}. Beyond the general role assignment, we further incorporate specialized mechanisms to enhance the accuracy of agent responses during generation. During adjacency planning, we design a looped verification mechanism between \textit{Adjacency} and \textit{Check}, forming a generate–verify–revise cycle that enforces spatial alignment under constraints. To ensure agents consistently fulfill their designated roles, we employ prompt engineering strategies: \textbf{Role-Play} and \textbf{Few-Shot Prompting} reinforce behavioral boundaries and contextual grounding, while the \textit{door\_window} agent uniquely adopts a \textbf{Chain-of-Thought (CoT)} strategy to reason from shared wall analysis to opening placement and style description. These strategies collectively enhance the controllability and semantic coherence of the agent-based chaining framework.

\begin{table}[t]
\small
\renewcommand{\arraystretch}{1.15}
\setlength{\tabcolsep}{4pt}
\begin{tabularx}{\linewidth}{p{2.2cm}X}
\toprule
\textbf{Agent} & \textbf{Duties} \\
\midrule
\textit{Manager} & Choose wall-structure or column-structure according to user input. \\
\hline
\textit{Allocation} & Wall: assign number and functions of rooms; Column: define grid and room count. \\
\hline
\textit{Adjacency} & Wall: Define spatial adjacency between rooms; Column: Assign rooms to occupied grid cells. \\
\hline
\textit{Check} & Validate adjacency logic and return to Adjacency if constraints are violated(Wall only). \\
\hline
\textit{Shape} & Generate room sizes and determine whether to add arc walls (Wall only). \\
\hline
\textit{Material} & Wall: Select materials for floors and walls; Column: Select materials for floors, columns, and beams. \\
\hline
\textit{Door\_Window} &  Wall: Select walls, plan opening sizes, and describe door/window styles; Column: Describe door/window styles. \\
\hline
\textit{Object} & Wall: Select floor (stable and relative) and wall objects; Column: Select floor (stable and relative) objects. \\
\bottomrule
\end{tabularx}
\caption{Agent roles and their duties (Wall for wall-structure scene and Column for column-structure scene).}
\label{tab:agent_roles}
\end{table}

\noindent\textbf{Agent Turn Control.} Our agent-based chaining framework follows the film set design pipeline in a strictly sequential manner. To regulate the speaking order of agents, we implement a Finite State Machine (FSM) \cite{wu2023autogen} with a state transition dictionary (\texttt{speaker\_transitions\_dict}) specifying the valid next speaker(s) for each agent. Moreover, each agent declares its context and speaking conditions through a natural language \texttt{description} field (e.g., ``I can only speak after \textit{Adjacency}, and the next speaker is \textit{Shape}''). These semantic constraints, together with FSM, provide a robust control mechanism for stable and predictable agent responses.

\noindent\textbf{Hooks for Data Bridging.} To seamlessly connect agent responses with the procedural generation pipeline, we adopt the hook mechanism from AutoGen~\cite{wu2024autogen}, which extracts agent response and converts it into standardized structured parameters in real time. For key agents, custom hooks continuously monitor responses and extract valid parameters that are directly forwarded to downstream generation functions. Beyond internal parameter passing, hooks also trigger external retrieval routines (including material selection, door/window asset matching, and object retrieval from SetDepot-Pro), thereby providing a unified interface that grounds high-level language descriptions into concrete generation operations for automated scene construction.

\section{SetDepot-Pro}

\begin{figure*}[htbp]
  \centering
  \includegraphics[width=1\textwidth]{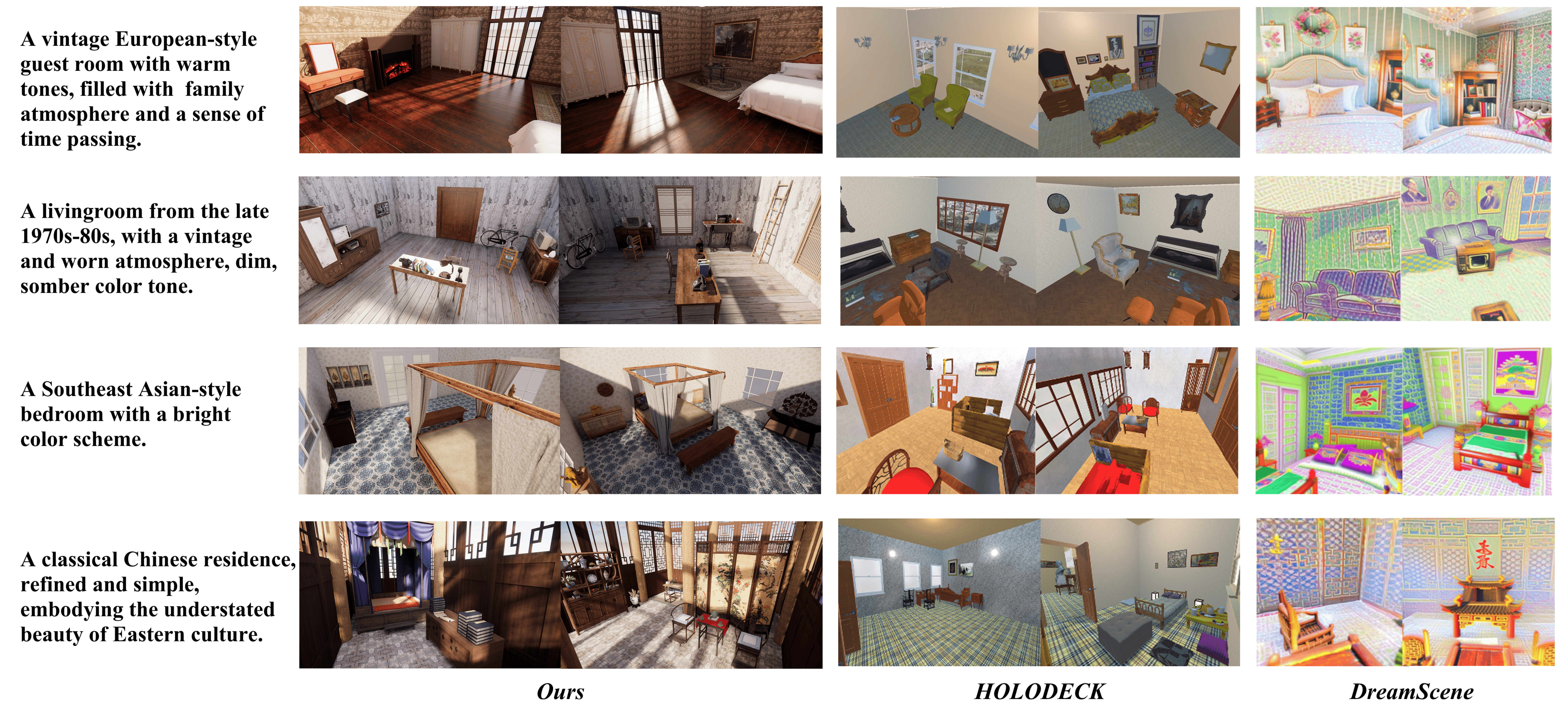}
  \caption{Qualitative comparison: A visual comparison of film scenes generated by our method, HOLODECK, and DreamScene. The results show that our method achieves stronger expressiveness in film scene generation, enabling scenes with distinct era, style, and regional characteristics, thereby ensuring greater historical and cinematic authenticity.}
  \label{fig:example}
\end{figure*}

\noindent\textbf{Collection and Pre-processing.}
The material library covers 733 materials across 12 categories (e.g., aged concrete, brick wall, flooring, marble, wallpaper, glass), sourced from professional repositories. The object library contains 6,862 assets obtained through three primary channels: (1) acquisitions from architectural studios, (2) purchases from model platforms, and (3) contributions from film art departments, all secured with appropriate usage rights. To enhance diversity, we collected not only general-purpose models (e.g., furniture for kitchens, offices, or restaurants) but also culturally distinctive items (e.g., Persian carpets, farming tools, ancient statues) and period-specific props from film sets and traditional Chinese palaces. For composite items such as table–chair ensembles, we retained them as unified assets to preserve contextual integrity. All assets underwent manual pre-processing, including resetting orientation, smoothing redundant geometry and scaling to real-world dimensions to ensure accurate spatial integration.

\noindent\textbf{Image Capture and Annotation.}
Standardized images were captured for each asset using automated camera controls in SketchUp, ensuring complete framing and consistent viewpoints. Object images were taken after orientation normalization, using the \texttt{active\_view} and \texttt{zoom\_extents} functions for consistency.  
Annotations were generated with GPT-4V, covering attributes such as category, style, cultural origin, and era. To ensure reliability, domain experts from film art departments reviewed critical labels, particularly those involving historical and cultural attributes. Additionally, materials with patterned textures were manually annotated with scale factors to preserve realistic proportions.

\noindent\textbf{Feature Encoding.}
After annotation, all labels were concatenated into unified textual descriptions for each asset. To enable semantic retrieval, we employed Sentence-BERT (all-mpnet-base-v2) to encode these descriptions into dense vector representations. The resulting features were normalized and stored in a precomputed embedding database for retrieval tasks. For the object library, features of door and window models were extracted and maintained as an independent category for door and window placement.

\begin{table*}[th]
    \centering
    \small
    \begin{tabular}{l c c c c c c}
        \toprule
        Method & Room Type & Layout\textuparrow & Material\textuparrow & Style\textuparrow & Object\textuparrow & Atmosphere\textuparrow \\  
        \midrule
        \multirow{4}{*}{HOLODECK} &  western guestroom & $ 7.2 \pm {\scriptstyle 0.4}$ & $6.0 \pm {\scriptstyle 0.0}$ &$8.0 \pm {\scriptstyle 0.0}$  & $7.0 \pm {\scriptstyle 0.0}$ & $7.4 \pm {\scriptstyle 0.49}$ \\
        & vintage room &$6.0 \pm {\scriptstyle 0.0}$  & $5.4 \pm {\scriptstyle 0.8}$ & $6.8 \pm {\scriptstyle 0.75}$ &$5.8 \pm {\scriptstyle 0.75}$  & $5.6 \pm {\scriptstyle 0.49}$ \\
        & asian-style bedroom &$7.0 \pm {\scriptstyle 0.0}$  &$6.0 \pm {\scriptstyle 0.0}$  &$8.0 \pm {\scriptstyle 0.0}$  & $7.2 \pm {\scriptstyle 0.4}$ & $6.8 \pm {\scriptstyle 0.4}$ \\
        & chinese residence &$5.4 \pm {\scriptstyle 0.49}$  & $4.4 \pm {\scriptstyle 0.49}$  & $3.4 \pm {\scriptstyle 0.49}$ &$3.4 \pm {\scriptstyle 0.49}$  & $4.4 \pm {\scriptstyle 0.49}$ \\
        \midrule
        \multirow{4}{*}{DreamScene} & western guestroom & $8.2 \pm {\scriptstyle 0.4}$  & $8.2 \pm {\scriptstyle 0.4}$  & $9.0 \pm {\scriptstyle 0.0}$  & $8.0 \pm {\scriptstyle 0.0}$  & $9.0 \pm {\scriptstyle 0.0}$  \\
        & vintage room & $6.4 \pm {\scriptstyle 0.49}$  & $5.4 \pm {\scriptstyle 0.49}$  & $7.8 \pm {\scriptstyle 0.4}$  & $6.8 \pm {\scriptstyle 0.4}$  & $6.2 \pm {\scriptstyle 0.75}$ \\
        & asian-style bedroom & $7.0 \pm {\scriptstyle 0.0}$ & $6.8 \pm {\scriptstyle 0.98}$ & $8.6 \pm {\scriptstyle 0.49}$ & $7.8 \pm {\scriptstyle 0.4}$ & $7.8 \pm {\scriptstyle 0.98}$ \\
        & chinese residence & $7.0 \pm {\scriptstyle 0.0}$  & $6.4 \pm {\scriptstyle 0.8}$  & $8.2 \pm {\scriptstyle 0.4}$  & $7.6 \pm {\scriptstyle 0.49}$  & $7.6 \pm {\scriptstyle 0.49}$ \\
        \midrule
        \multirow{4}{*}{Ours} & western guestroom & $8.2 \pm {\scriptstyle 0.4}$  & $8.4 \pm {\scriptstyle 0.49}$  & $9.0 \pm {\scriptstyle 0.0}$  & $8.2 \pm {\scriptstyle 0.4}$  & $9.0 \pm {\scriptstyle 0.4}$  \\
        & vintage room & $7.4 \pm {\scriptstyle 0.9}$  & $8.2 \pm {\scriptstyle 0.4}$  & $9.0 \pm {\scriptstyle 0.0}$  & $8.2 \pm {\scriptstyle 0.4}$  & $8.6 \pm {\scriptstyle 0.49}$ \\
        & asian-style bedroom & $7.6 \pm {\scriptstyle 0.49}$ & $7.8 \pm {\scriptstyle 0.4}$ & $9.0 \pm {\scriptstyle 0.0}$ & $7.8 \pm {\scriptstyle 0.4}$ & $8.6 \pm {\scriptstyle 0.49}$ \\
        & chinese residence & $8.4 \pm {\scriptstyle 0.49}$  & $8.6 \pm {\scriptstyle 0.49}$  & $9.0 \pm {\scriptstyle 0.0}$  & $8.2 \pm {\scriptstyle 0.4}$  & $9.0 \pm {\scriptstyle 0.0}$ \\
        \bottomrule
    \end{tabular}
    \caption{Quantitative results: A comparison among our method, HOLODECK, and DreamScene across five key evaluation metrics: Object Layout, Material Selection, Style Consistency, Object Selection, and Overall Aesthetic Atmosphere.}
    \label{table:comparison}
\end{table*}

\section{Experiment}

\noindent\textbf{Implementation Details.}
For the LLM, we employ OpenAI’s GPT-4o \cite{hurst2024gpt} with a temperature setting of 0.7. For semantic retrieval, the Sentence-BERT model employed is all-mpnet-base-v2 \cite{reimers2019sentence}. All reported results were obtained on a MacBook equipped with an Apple M2 chip and 16GB of memory. Procedural generation was carried out within SketchUp 2024 \cite{sketchup}, and the rendered visualizations were produced using Enscape \cite{enscape}.

\noindent\textbf{Metric.} Following prior work~\cite{ccelen2024design,wu2024gpt} demonstrating GPT-4V’s alignment with human judgments in 3D evaluation, we adopt a GPT-4V-based criteria to evaluate the quality of scenes generated by our method and the baseline works (HOLODECK \cite{yang2024holodeck} and DreamScene \cite{li2024dreamscene}), assessing their suitability for cinematic use. The evaluation covers five core aspects: \textit{Object Layout}, \textit{Material Selection}, \textit{Style Consistency}, \textit{Object Selection}, and \textit{Overall Aesthetic Atmosphere}. Specifically, Object Layout examines spatial balance and composition; Material Selection evaluates whether the chosen materials match the style described in the natural language description; Style Consistency checks alignment with historical or thematic intent; Object Selection evaluates whether the chosen objects align with the natural language description and maintain cinematic realism; Overall Aesthetic Atmosphere evaluates the overall harmony among material, door and window, and object selection within the scene. Each aspect is rated on a scale of 0-10, with higher scores indicating stronger cinematic alignment. To ensure robustness, each scene is scored 10 times independently (by setting $n=10$ in the GPT-4V API), and we report the average score and standard deviation for each metric.

\noindent\textbf{Quantitative Analysis.} As shown in Table~\ref{table:comparison}, our method consistently achieves higher scores in layout design, material realism, style consistency, object selection, and overall aesthetic atmosphere. The \textit{chinese residence} setting demonstrates the most notable improvements. Our method achieves scores of 8.4 in layout, 8.6 in material, and 9.0 in style, substantially outperforming HOLODECK (5.4, 4.4, 3.4) and DreamScene (7.0, 6.4, 8.2) on the same metrics. These gains are attributed to our adoption of a culturally accurate column-structure scene, which better reflects the historical and architectural semantics of traditional Chinese design. In other scenes, our method continues to outperform. For the \textit{vintage room}, it achieves 7.4 in layout and 8.6 in atmosphere, exceeding DreamScene (6.4, 6.2) and HOLODECK (6.0, 5.6). In the \textit{asian-style bedroom}, it leads with 7.6 and 8.6. These consistent advantages underscore our superior spatial planning and aesthetic control. Overall, the results demonstrate our framework's capability to generate semantically grounded and stylistically faithful film sets across diverse cinematic contexts. We also evaluate HOLODECK on SetDepot-Pro (see supplementary material).

\noindent\textbf{Qualitative Analysis.} Figure~\ref{fig:example} presents a visual comparison among our method, HOLODECK, and DreamScene across 4 scene types under identical prompts. In the European-style room, our design captures warm tones and temporal depth, while HOLODECK loses structural integrity and DreamScene over-saturates colors. The 1980s livingroom generated by our method retains authentic wear and dim tones, whereas HOLODECK fails to reflect aging, and DreamScene appears visually inconsistent. For the Southeast Asian bedroom, our method incorporates regionally appropriate furniture and decorative patterns. DreamScene, while visually rich, suffers from cluttered layout, and HOLODECK lacks tropical characteristics. The classical Chinese residence best highlights our structural advantage: we adopt column-structure and traditional furnishings aligned with historical aesthetics, in contrast to the wall-structure and stylistically mismatched outputs of HOLODECK and DreamScene. These results highlight our strength in cultural specificity and cinematic coherence.

\noindent\textbf{Ablation Study.} In our framework, the \textit{Adjacency} agent proposes room-to-room relationships, while the \textit{Check} agent validates and refines them. We compare \textit{Adjacency} alone with \textit{Adjacency + Check} across layouts of 3–5 rooms. As room count grows, \textit{Adjacency} alone frequently produces invalid connections, whereas incorporating \textit{Check} systematically corrects errors. Over 10 samples per setting, the combined setup achieves higher correctness rates (Table~\ref{tab:adjacency_results}), particularly in more complex room adjacency.

To evaluate the impact of different prompting strategies, we conducted an ablation study on three key agents as shown in Table~\ref{tab:prompt_strategy_ablation}. For each agent, we defined three evaluation criteria and scored three samples per setting on a scale of 1-5. The scores were summed across the three criteria (maximum 15) and then normalized to the range $[0,1]$. Final results were averaged across all samples. Detailed scoring criteria are provided in the supplementary material. All prompting strategies improved performance, with Few-Shot prompting showing the most consistent gains, and Chain of Thought proving particularly effective for \textit{Door\_Window}.

\begin{table}[t]
    \centering
    \begin{tabular}{l c c c c}
        \toprule
        Strategy & 3 Rooms & 4 Rooms & 5 Rooms  \\
        \midrule
        Adjacency  & 90\% & 50\% & 30\% \\
        Adjacency + Check & 100\% & 60\% & 50\%\\
        \bottomrule
    \end{tabular}
    \caption{Ablation study on \textit{Adjacency} and \textit{Check}, conducted to evaluate the impact of \textit{Check} on improving the accuracy of parameters generated by \textit{Adjacency}.}
    \label{tab:adjacency_results}
\end{table}

\begin{table}[ht]
\centering
\small
\setlength{\tabcolsep}{6pt}
\begin{tabular}{l|cccc}
\toprule
Agent & Prompt & +Role-Play & +Few-Shot & +CoT \\
\midrule
Material     & 0.77 & 0.80 & 0.89 & / \\
Door\_Window   & 0.73 & 0.82 & 0.87 & 0.95 \\
Object          & 0.75 & 0.80 & 0.93 & / \\
\bottomrule
\end{tabular}
\caption{Normalized scores for each agent under different prompting strategies. ``/'' indicates that the strategy was not applied, as these tasks did not require multi-step reasoning.}
\label{tab:prompt_strategy_ablation}
\end{table}

\noindent\textbf{User Study.} To evaluate the effectiveness of our method in realistic film set design scenario, we conducted a user study involving 32 film industry professionals. The study covered 4 representative scene types, each generated by our method, HOLODECK, and DreamScene. Participants rated the results across five key criteria: text-structure consistency, material selection, object selection, aesthetic atmosphere, and spatial composition. As summarized in Table~\ref{tab:user_study}, our method achieved the highest average scores across all evaluation dimensions, indicating superior structural coherence, visual realism, and alignment with cinematic design goals.

\begin{table}[ht]
\centering
\small
\setlength{\tabcolsep}{4pt}
\begin{tabular}{l|ccccc}
\toprule
Method & Struc. & Mat. & Obj. & Atmos. & Spa. \\
\midrule
HOLODECK     & 2.59 & 2.67 & 2.74 & 2.66 & 2.85 \\
DreamScene   & 2.71 & 2.34 & 2.41 & 2.21 & 2.41 \\
Ours          & 4.88 & 4.62 & 4.66 & 4.55 & 4.53 \\
\bottomrule
\end{tabular}
\caption{User study results across five evaluation criteria: structural coherence, material selection, object selection, aesthetic atmosphere, and spatial composition. Scores are on a scale of 1-5, with the higher the better.}
\label{tab:user_study}
\end{table}

\section{Conclusion}
We presented FilmSceneDesigner, a novel scene generation system that emulates professional film set design workflow. By combining an agent-based chaining framework with procedural generation, our method enables dynamic parameter selection, ensuring greater flexibility in scene composition. To further enhance cinematic authenticity, we constructed SetDepot-Pro, a film-specific dataset spanning different historical periods and architectural styles, allowing more contextually accurate scene generation. Experimental results demonstrate that our approach effectively produces cinematically authentic scenes that meet industry standards of film set design. User studies further validate that our system generates visually coherent and structurally realistic environments, making it a valuable tool for film set design.

\section{Acknowledgments}
This work is supported by the Natural Science Foundation of Shanghai (Grant No. 25ZR1401130 and No. 24ZR1422400), the National Natural Science Foundation of China (Grant No. 62402306),  and the Open Research Project of the State Key Laboratory of Industrial Control Technology, China (Grant No. ICT2024B72).

\bibliography{aaai2026}

\clearpage
\appendix

\twocolumn[
\begin{center}
{\LARGE \bfseries
FilmSceneDesigner: Chaining Set Design for Procedural Film Scene Generation
}

\vspace{0.8em}

{\Large \bfseries
Supplementary Material
}
\end{center}

\vspace{2em}
]

\section{Details of SetDepot-Pro}
To address the lack of film-specific digital assets in existing datasets, we construct \textbf{SetDepot-Pro}, a depot of film set assets designed for professionals, which includes 6,862 high-quality object assets and 733 materials tailored for cinematic scene creation.

\begin{figure}[h]
  \centering
  \includegraphics[width=\linewidth]{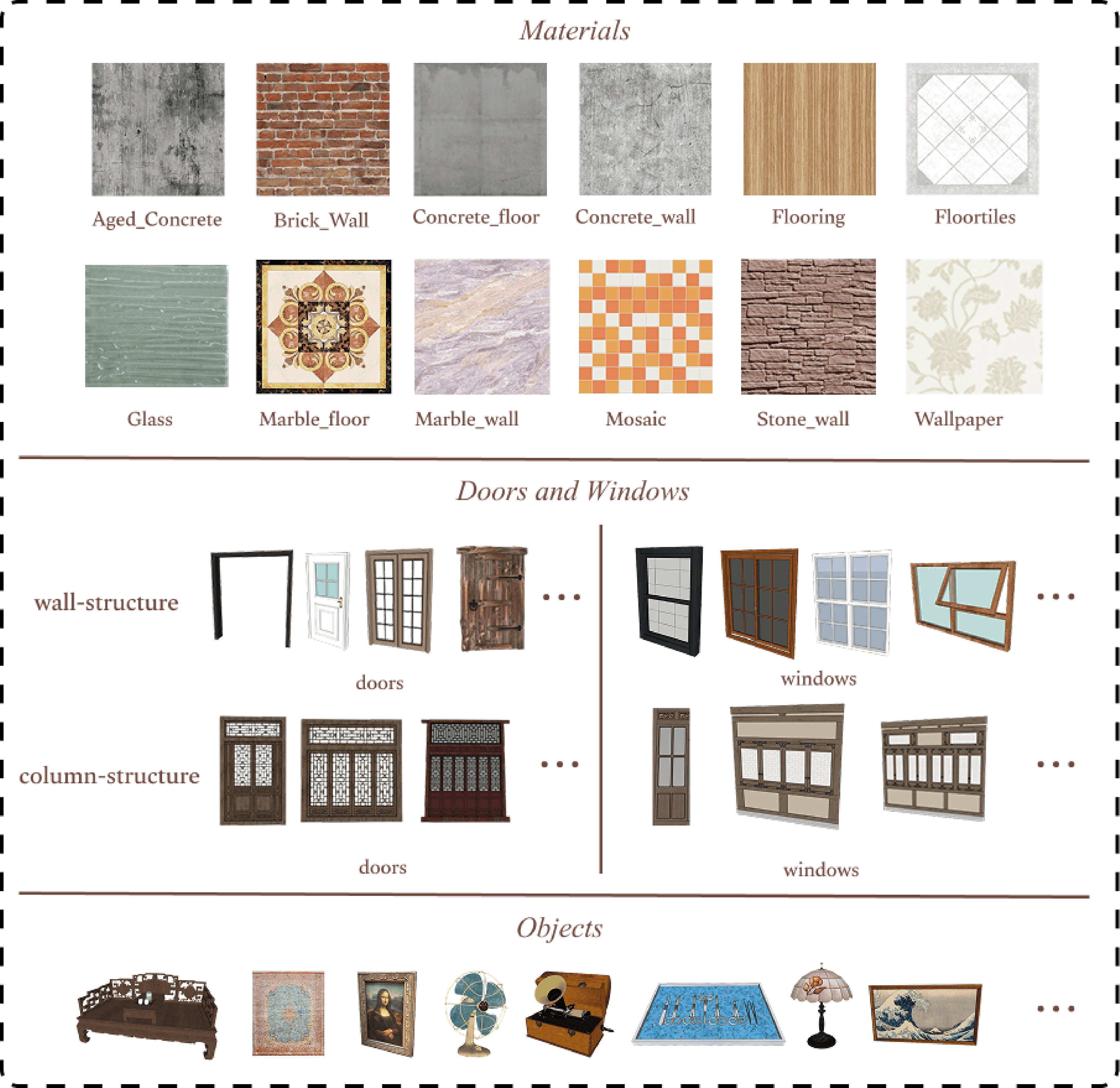}
  \caption{Categories and representative examples of assets in SetDepot-Pro, including materials (12 categories), doors and windows for both wall-structure and column-structure scenes, and diverse objects covering a wide range of scene types and cultural regions.}
  \label{fig:dataset_cat}
\end{figure}

\noindent\textbf{Categorization.} SetDepot-Pro contains both materials and objects tailored for film set design, as shown in Figure~\ref{fig:dataset_cat}. For the material library, we collect 733 materials across 12 categories from material websites, including aged concrete, brick wall, concrete floor, concrete wall, flooring, floor tiles, glass, marble floor, marble wall, mosaic, stone wall, and wallpaper. Within the object library, we extract all door and window assets into a dedicated subset with separately computed features to support the door and window placement stage and enhance retrieval accuracy. These assets include a variety of doors (e.g., door frames, single, double, and sliding doors) and windows (e.g., fixed and sliding types) for wall-structure scenes, as well as traditional Chinese lattice-style doors and windows with variations in panel counts and lengths for column-structure scenes. Beyond doors and windows, the object library also contains a wide range of models covering diverse scene types such as Japanese courtyards, hospitals and clinics, convenience stores, tea restaurants, classrooms, and prisons, among others. To enrich cultural and regional diversity, we further incorporate distinctive assets from architectural studios, including Persian carpets, rural farming tools, ancient statues, and models representing styles from Egypt, Pakistan, Byzantium, the Mediterranean, Southeast Asia, Russia, China’s Xinjiang and Tibet regions, Mongolia, Morocco, Nepal, Turkey, and India. Unlike the material library, we do not impose rigid categorization on these object assets; instead, retrieval relies on their textual descriptions, which enables more flexible and semantic indexing.

\begin{figure}[h]
  \centering
  \includegraphics[width=\linewidth]{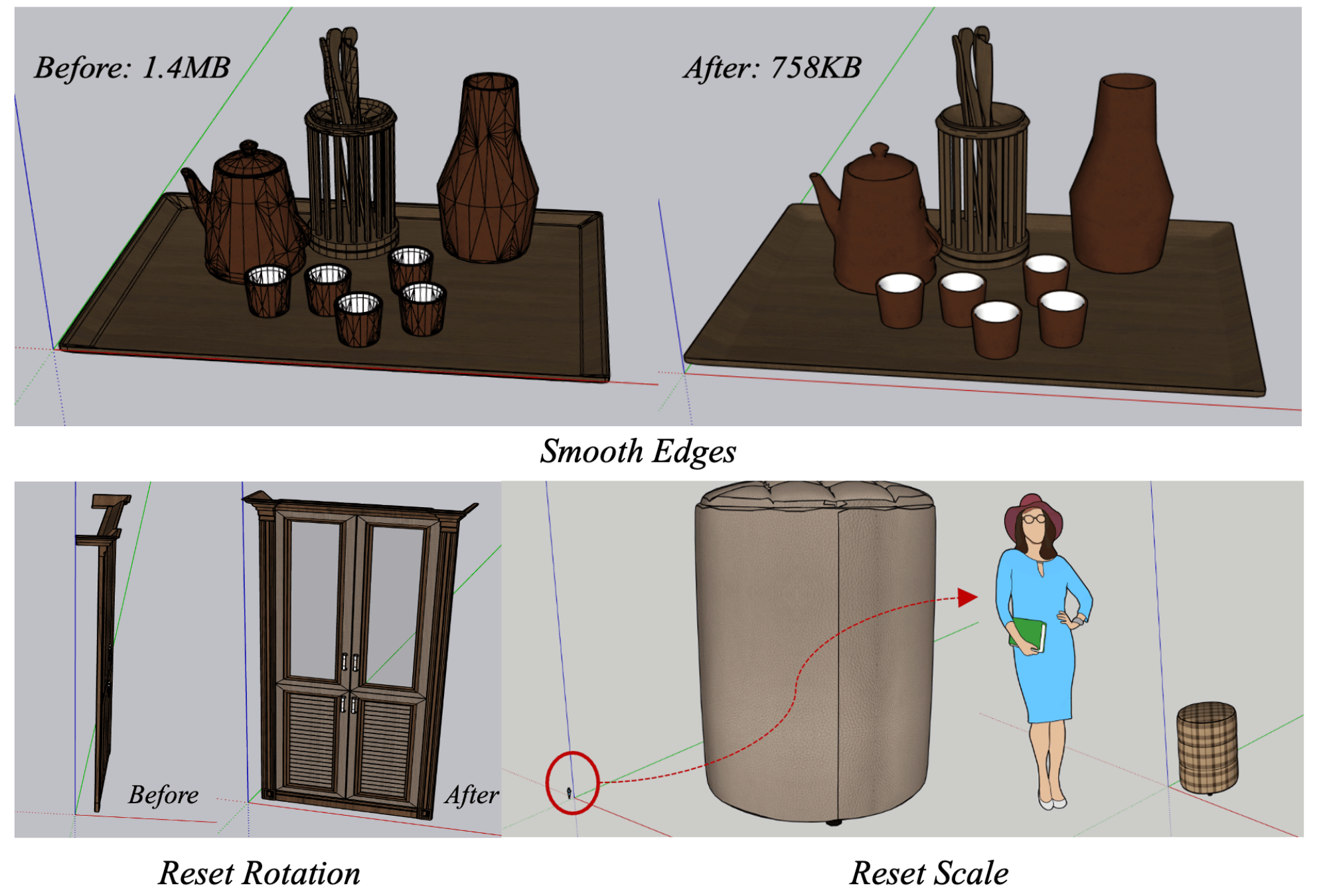}
  \caption{Preprocessing steps: Smooth Edges, Reset Rotation, and Reset Scale.}
  \label{fig:model_preprocess}
\end{figure}

\noindent\textbf{Pre-processing.}
After collecting the assets, we preprocess the collected object and door/window models. As shown in Figure~\ref{fig:model_preprocess}, this preprocessing consists of three steps: Smooth Edges, Reset Rotation, and Reset Scale. First, due to SketchUp’s performance limits, handling models with excessive faces and edges incurs high computational costs. We apply Smooth Edges to reduce complexity and ensure stable retrieval and placement of multiple objects in a scene, which significantly decreases file size (e.g., from 1.4MB to 758KB). Second, since our models come from diverse sources with varying orientations, we apply Reset Rotation to facilitate accurate alignment during the layout stage. Specifically, we define the model’s front view on the plane formed by SketchUp’s red and blue axes, observed from the positive green axis. This ensures consistent orientation, such as the correctly aligned door model. Finally, we apply Reset Scale to standardize object dimensions to real-world scale by adjusting the object’s bounding box. For example, red circles indicate human figures of the same height, demonstrating how we scale objects like chairs to match realistic proportions. After applying these preprocessing operations, the total size of SetDepot-Pro is reduced from 68.64GB to 33.85GB, and all object proportions and views are standardized.

\begin{figure}[h]
  \centering
  \includegraphics[width=\linewidth]{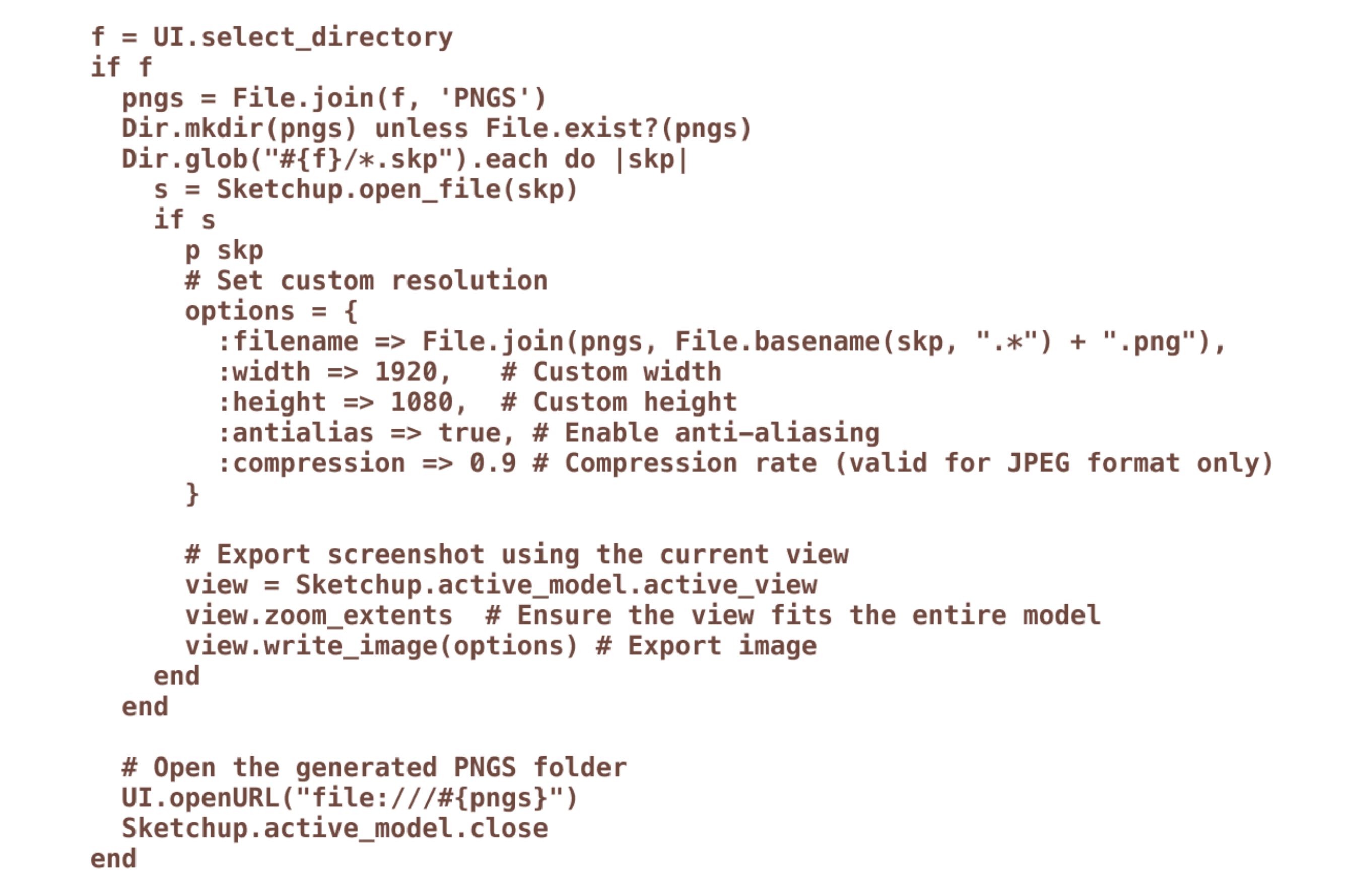}
  \caption{Implementation for capturing images of object and door/window assets in SetDepot-Pro.}
  \label{fig:capture_dataset}
\end{figure}

\begin{figure}[h]
  \centering
  \includegraphics[width=\linewidth]{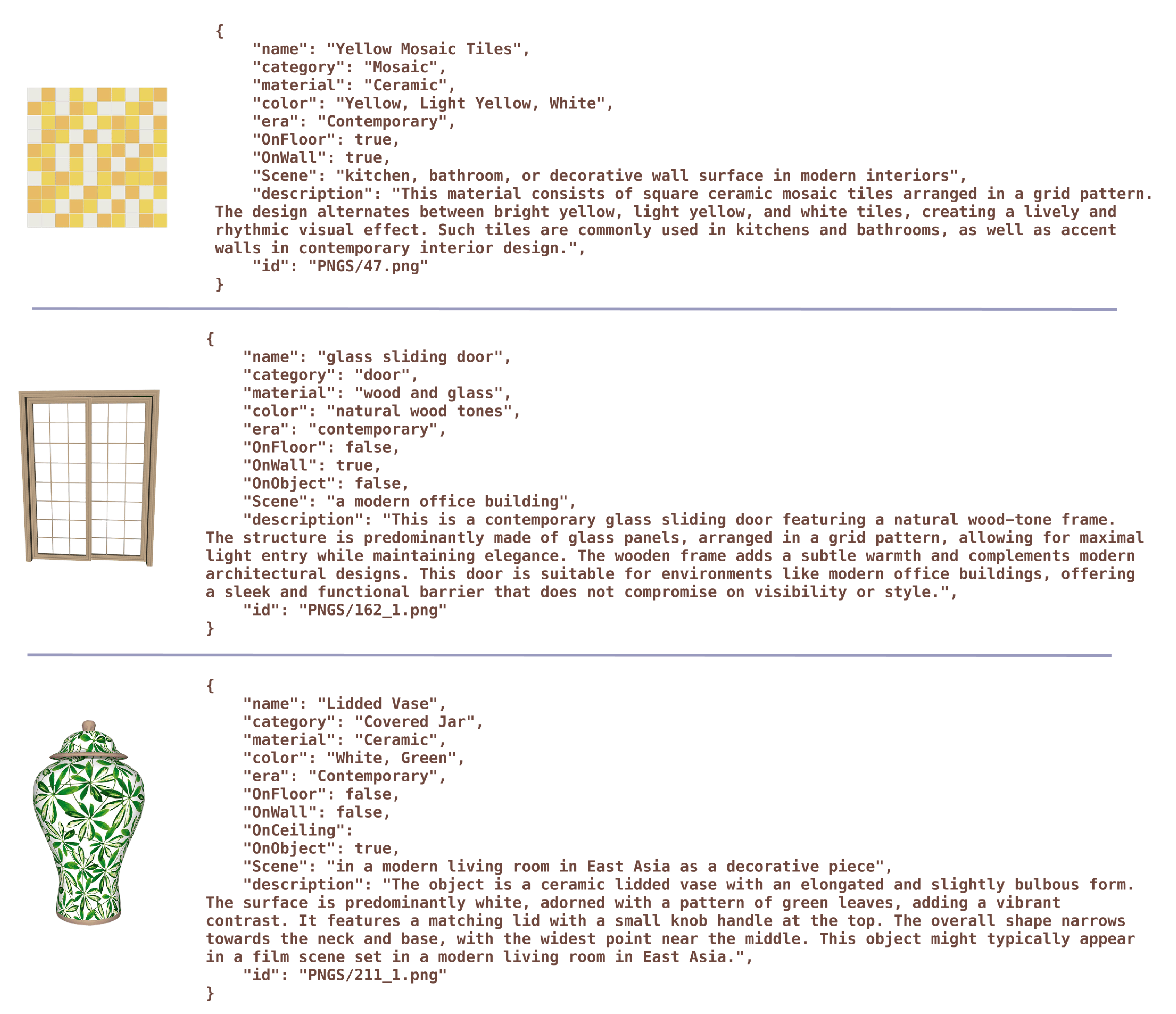}
  \caption{Examples of annotated assets in SetDepot-Pro, including a material, a door, and an object. Each asset is paired with its corresponding annotation, containing attributes such as category, material, color, era, applicable placement, scene description, and unique ID.}
  \label{fig:labels}
\end{figure}

\noindent\textbf{Image Capture and Annotation.}
In this stage, we capture images for both materials and objects, and subsequently perform annotation using GPT-4V. For the material library, we collect each material in two formats: PNG and SKP. The PNG format is utilized for annotation purposes, while the SKP format is used for material assignment on walls and floors, as it ensures more realistic results. For the object library, we reset the rotation of each model during preprocessing and capture images based on a standardized front view. The code for capturing object images is shown in Figure~\ref{fig:capture_dataset}. This ensures that the exported image fully displays the entire object and avoids unnecessary blank areas around it. After capturing the images, we use GPT-4V to generate annotation labels, as shown in Figure~\ref{fig:labels}.

\section{Procedural Generation Functions}
We implement a set of dedicated functions to enable procedural scene construction in SketchUp, performing generation tasks across four stages: \textit{floorplan and structure generation}, \textit{material assignment}, \textit{door and window placement}, and \textit{object retrieval and layout}. These functions translate the parameters produced by our agent-based chaining framework into concrete geometric and semantic operations in SketchUp. Table~\ref{tab:function_responsibilities} summarizes the key functions and their responsibilities. The implementation of these functions is provided in the supplementary code file.

\begin{table*}[ht]
\centering
\small
\begin{tabular}{l l l}
\toprule
\textbf{Module} & \textbf{Function} & \textbf{Responsibility} \\
\midrule
\multirow{9}{*}{Structure} 
 & \texttt{parse\_edge} & Parse boundary edges from room definitions \\
 & \texttt{add\_line} & Create a straight edge between two points \\
 & \texttt{add\_arc\_by\_start\_end\_chord} & Create an arc from endpoints and chord height \\
 & \texttt{add\_face\_from\_edges} & Generate a planar face enclosed by edges \\
 & \texttt{offset} & Apply thickness and extrude geometry \\
 
 & \texttt{pushpull} & wall height \\
 & \texttt{add\_circle\_by\_center\_radius} & Create circular base for column placement \\
\midrule
\multirow{3}{*}{Material} 
 & \texttt{set\_attribute} & Assign labels such as \texttt{roomX\_idY}, \texttt{roomX\_floor} \\
 & \texttt{apply\_material} & Apply textures based on Attributes \\
\midrule
\multirow{4}{*}{Door\_Window} 
 & \texttt{open\_wall} & Create a door or window opening on target wall \\
 & \texttt{rotate\_model} & Align model orientation with wall direction \\
 & \texttt{scale\_model} & Resize model to match opening dimensions \\
 & \texttt{translate\_model} & Move model into final placement position \\
\midrule
\multirow{7}{*}{Object} 
 & \texttt{place\_corner\_objects} & Place anchor objects at corners of space \\
 & \texttt{place\_edge\_objects} & Place anchor objects along walls or edges \\
 & \texttt{place\_center\_object} & Center anchor object in the space \\
 & \texttt{place\_relative\_object} & Place object relative to anchor (e.g., left, right) \\
 & \texttt{place\_wall\_objects} & Place decorative objects on designated wall faces \\
\bottomrule
\end{tabular}
\caption{Key procedural generation functions and their responsibilities.}
\label{tab:function_responsibilities}
\end{table*}

\section{Prompts for Set Design Agents}
\begin{figure*}[h]
  \centering
  \includegraphics[width=1\textwidth]{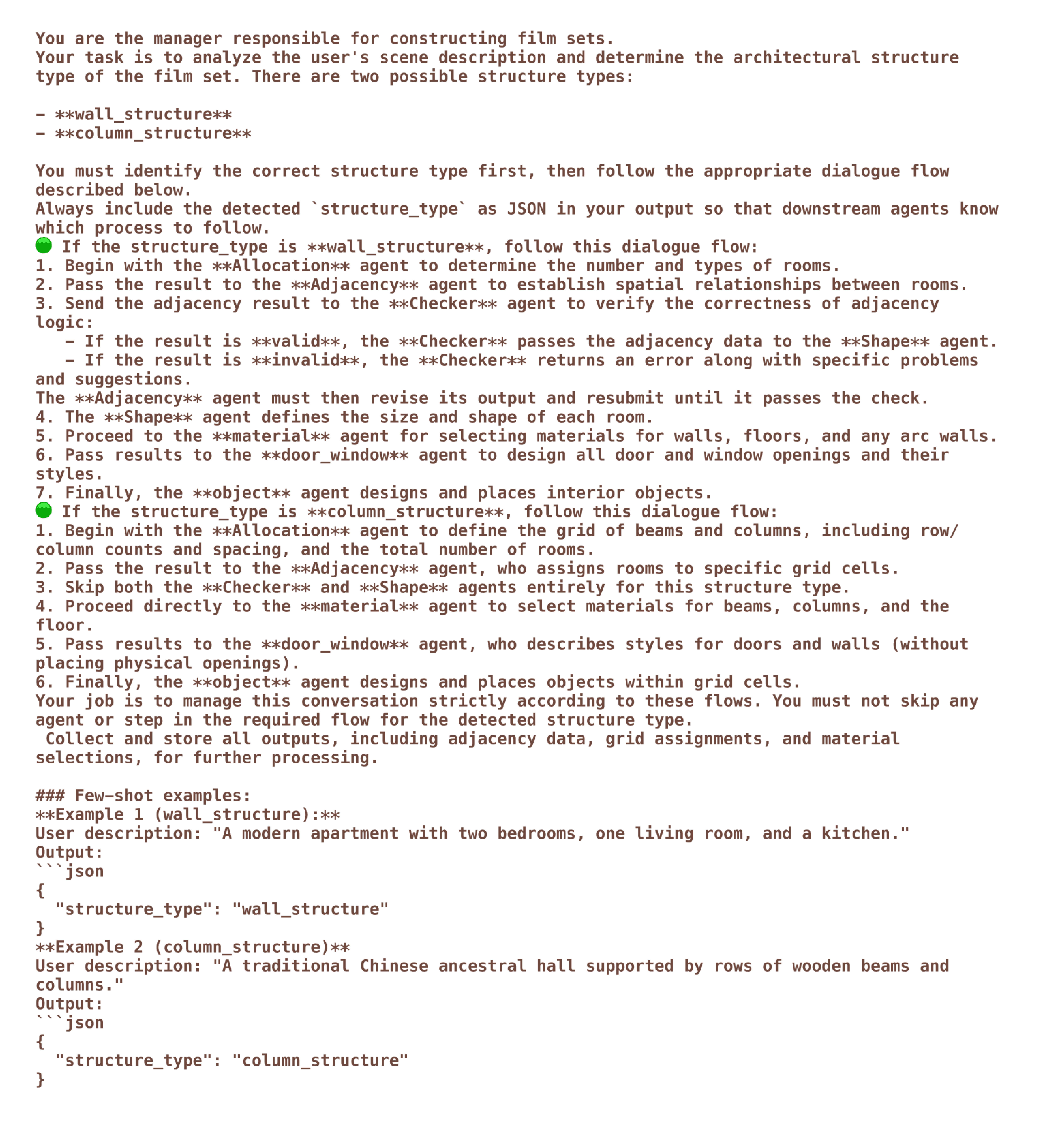}
  \caption{Prompts for the \textit{Manager} agent.}
  \label{fig:Manager}
\end{figure*}

\begin{figure*}[h]
  \centering
  \includegraphics[width=1\textwidth]{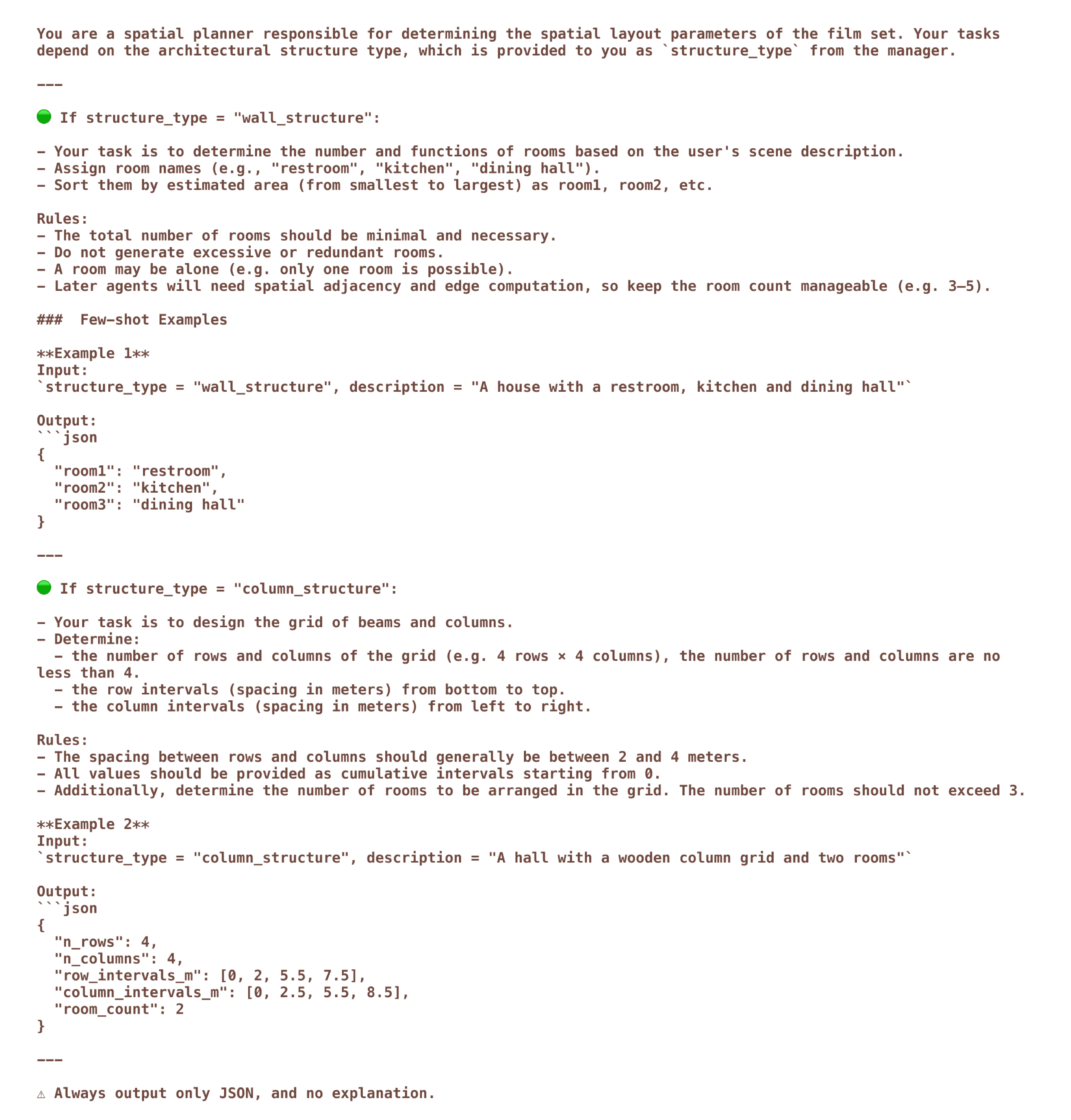}
  \caption{Prompts for the \textit{Allocation} agent.}
  \label{fig:Allocation}
\end{figure*}

\begin{figure*}[h]
  \centering
  \includegraphics[width=1\textwidth]{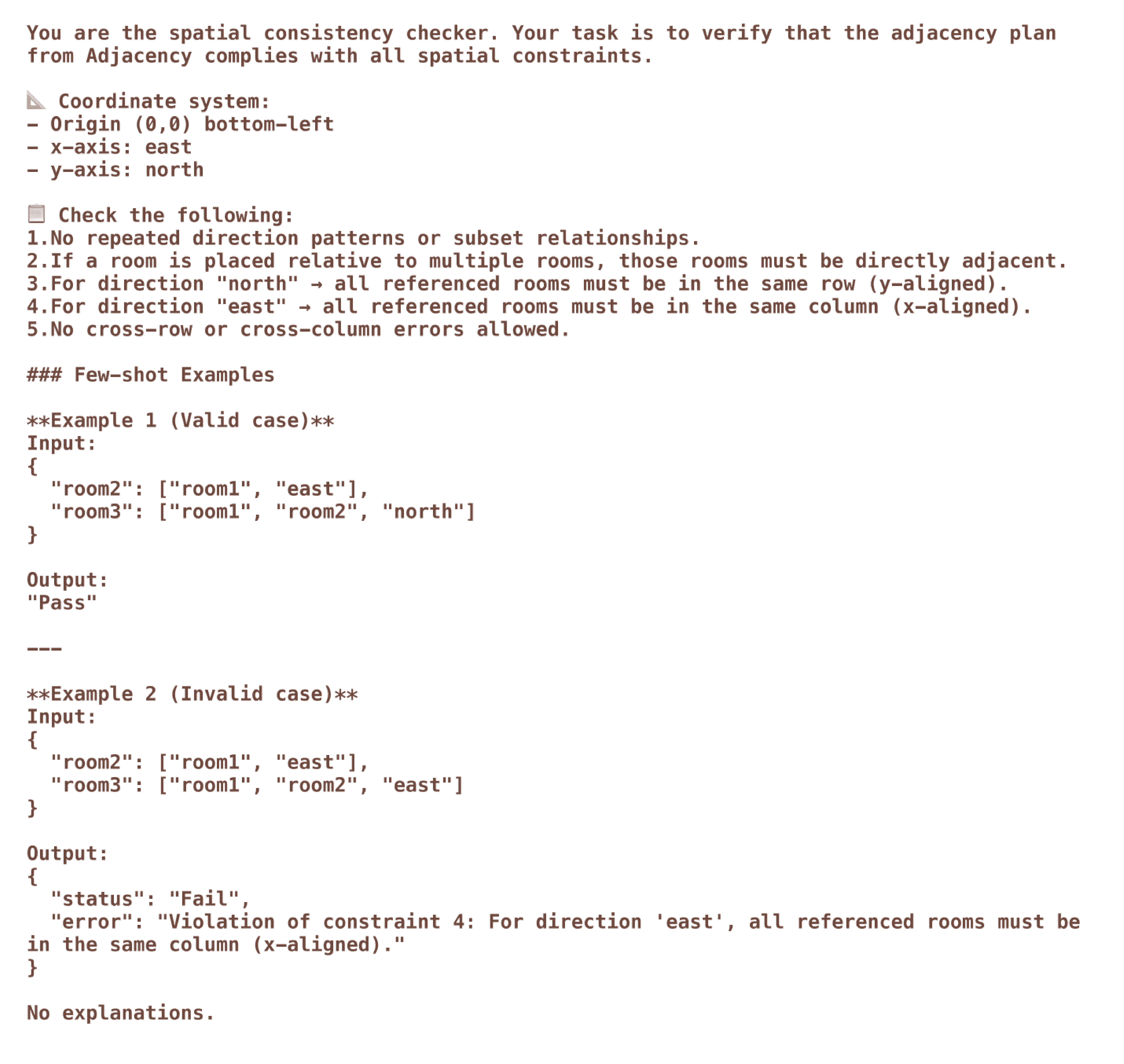}
  \caption{Prompts for the \textit{Check} agent.}
  \label{fig:Check}
\end{figure*}

We design an agent-based chaining framework composed of a series of set design agents, each responsible for parameter planning across the four generation stages. For completeness, we provide the detailed prompts that define the behavior of each agent in our framework. Each agent prompt specifies its role, responsibility, and reasoning scope during procedural scene generation. Figures~\ref{fig:Manager}--\ref{fig:Object_part2} illustrate these definitions for all agents, including \textit{Manager}, \textit{Allocation}, \textit{Adjacency}, \textit{Check}, \textit{Shape}, \textit{Material}, \textit{Door\_Window}, and \textit{Object}. For agents with longer prompts, such as \textit{Door\_Window} (Figures~\ref{fig:Door_Window_part1} and~\ref{fig:Door_Window_part2}) and \textit{Object} (Figures~\ref{fig:Object_part1} and~\ref{fig:Object_part2}), we split the content into two parts for clarity.

\section{Scoring Criteria of Ablation Study}
As shown in Table 4 of the main paper, we perform an ablation study on three key agents (Material, Door\_Window, and Object) to analyze how different prompting strategies affect the quality of parameter generation. For each agent, we define three agent-specific evaluation criteria, and the outputs are scored by GPT-4o. For each prompt strategy, we generate three test cases per agent and score each output across these dimensions. Each criterion is rated on a scale of 1–5, resulting in a maximum total score of 15 per case. The reported final score for each prompt strategy is the average of the three trials.
The specific evaluation questions for each agent are as follows:

\noindent\textbf{Material Agent}  
(1) Does the output follow the constraints? (1 = completely incorrect, 5 = fully correct)  
(2) Are the material descriptions consistent with the user input? (1 = completely mismatched, 5 = highly matched)  
(3) Are the material descriptions complete, covering type, color, texture, and style? (1 = incomplete, 5 = very complete)  

\noindent\textbf{Door\_Window Agent}  
(1) Does the agent correctly handle shared wall reasoning? (1 = completely incorrect, 5 = fully correct)  
(2) Are the door/window placements valid, reasonable, and sufficient in number? (1 = invalid or insufficient, 5 = highly valid and sufficient)  
(3) Does each opening include a detailed and natural model description? (1 = unclear, 5 = highly detailed and natural)  

\noindent\textbf{Object Agent}  
(1) Is the spatial relationship between anchor objects and relative objects valid and well-formed? (1 = completely invalid, 5 = highly valid)  
(2) Are the object descriptions clear, semantically accurate, and well-structured? (1 = inaccurate/unclear, 5 = very clear and accurate)  
(3) Are the parameters complete, including all required fields and values? (1 = severely incomplete, 5 = fully complete)

\section{Scoring Criteria of User Study}
As shown in Table 5 of the main paper, we conduct a user study to evaluate the effectiveness of our method in realistic film set design, inviting 32 film industry professionals to participate. In this study, participants rate each generated scene on five critical dimensions of cinematic set design. The specific evaluation criteria are as follows:

(1) Structural Coherence: Does the architectural layout of the scene align with the spatial structure described in the input text? (1 = completely inconsistent, 5 = perfectly consistent)

(2) Material Selection: Do the selected materials enhance the cinematic atmosphere and match the era, region, and style specified in the description? (1 = completely mismatched, 5 = highly matched)

(3) Object Selection: Are the props and furniture in the scene consistent with the narrative setting and do they contribute to visual authenticity? (1 = irrelevant, 5 = highly relevant)

(4) Aesthetic Atmosphere: Does the overall scene composition adhere to the aesthetic standards of film set design and provide an immersive experience? (1 = not immersive at all, 5 = highly immersive and aesthetically coherent)

(5) Spatial Composition: Does the scene exhibit sufficient spatial depth and layering to support camera placement and movement during filming? (1 = insufficient, 5 = highly sufficient)

\section{Analysis of Retrieval Strategy}
In our method, retrieval is performed solely based on semantic similarity, while scale compatibility is handled at a later placement stage through adaptive scaling. This design choice aims to preserve semantically optimal candidates during retrieval, which is particularly important for film-scene generation tasks where temporal and regional attributes play a critical role. To validate the effectiveness of this design under the SetDepot-Pro task setting, we analyze how different retrieval strategies respond to identical film-oriented object queries. Specifically, we select five object categories from the Objaverse dataset that exhibit strong temporal and regional characteristics as defined in SetDepot-Pro. For each query, the same textual description is used for retrieval under both strategies, and we compare the semantic similarity scores of the Top-1 retrieved assets. As shown in Table~\ref{tab:size_constraint_retrieval}, our retrieval strategy consistently yields higher Top-1 semantic similarity scores under identical queries. This result indicates that deferring scale handling to the placement stage helps retain semantically more relevant candidates during retrieval, thereby better supporting the semantic requirements of film-scene generation.

\begin{table}[t]
\centering
\begin{tabular}{l c c}
\toprule
Object Query & Ours $\uparrow$ & Holodeck $\uparrow$ \\
\midrule
1920s Armchair & 31.04 & 31.04 \\
Victorian Desk & 30.22 & 28.75 \\
1950s Sideboard & 30.64 & 30.64 \\
Edo-period Tea Table & 32.06 & 30.56 \\
French Rococo Console & 30.80 & 29.70 \\
\bottomrule
\end{tabular}
\caption{Comparison of retrieval strategies}
\label{tab:size_constraint_retrieval}
\end{table}

\section{Application}

\textbf{Multi-room Scene Generation and Editing.} Given the user input \textit{“A modern police station with a prison cell room, a security room, and an officer's office, featuring a minimalist, functional style and a serious, professional atmosphere”}, our system automatically generates a complete scene with three rooms at once, providing a ready-to-use multi-room set design, as shown in Figure~\ref{fig:edit}. The whole generation process is further demonstrated in the supplementary demo video.

After the full scene has been procedurally constructed, FilmSceneDesigner enables users to perform interactive editing operations directly within the SketchUp interface, as follows:  

(1) \textbf{Add doors or windows}: Select a target wall face, specify the opening size, and choose a door or window model. The system creates the opening on the wall and sequentially performs model loading, rotation, scaling, and translation to ensure the new component is properly aligned and attached to the wall surface.  

(2) \textbf{Change existing doors or windows}: Select the original component and specify a new model. The system automatically inherits the direction and size attributes of the associated wall, applying the appropriate geometric transformations to ensure the new model is spatially aligned with the previous one.  

(3) \textbf{Change wall or floor materials}: Select any face on the floor, interior wall, or exterior wall and upload a custom material file. The system applies the new material in batch mode based on the face’s attribute (wall id or floor).  

(4) \textbf{Change the material of individual objects}: Select a specific object, where the system recursively extracts all materials used within the component and presents them in a graphical interface. Users can then override the target material by specifying a new color or uploading a texture image.  

(5) \textbf{Change objects in the scene}: Substitute any selected object with a new model. The system captures the anchor point (bottom-center) and rotation angle of the original component and applies the same transformation to the new model, enabling seamless one-to-one replacement at the individual object level.  

\begin{figure}[h]
  \centering
  \includegraphics[width=\linewidth]{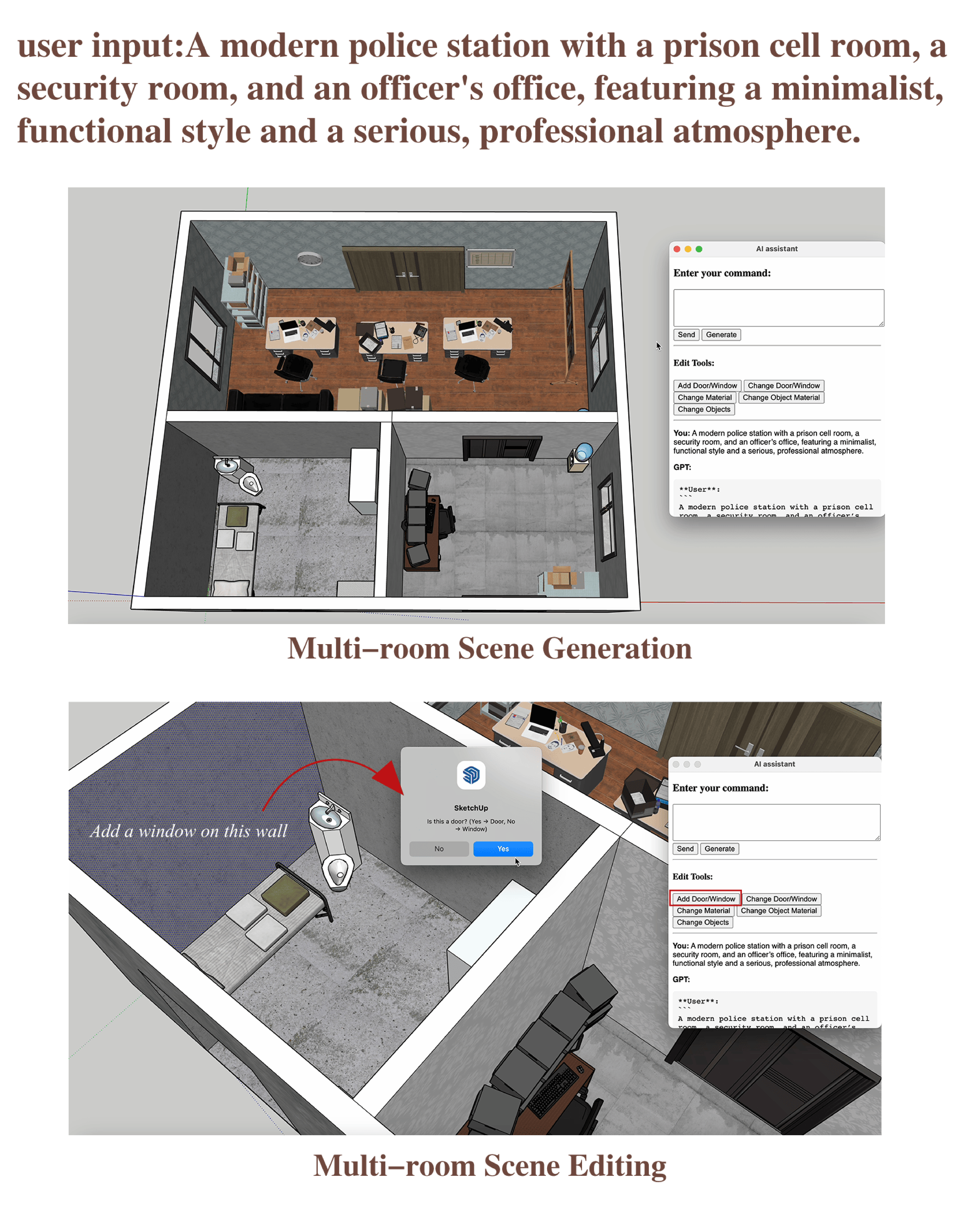}
  \caption{Multi-room scene generation and editing in FilmSceneDesigner. Given a natural language input, the system procedurally generates a complete three-room scene (top). After generation, users can perform interactive editing directly in SketchUp.}
  \label{fig:edit}
\end{figure}

\begin{figure}[t]
  \centering
  \includegraphics[width=\linewidth]{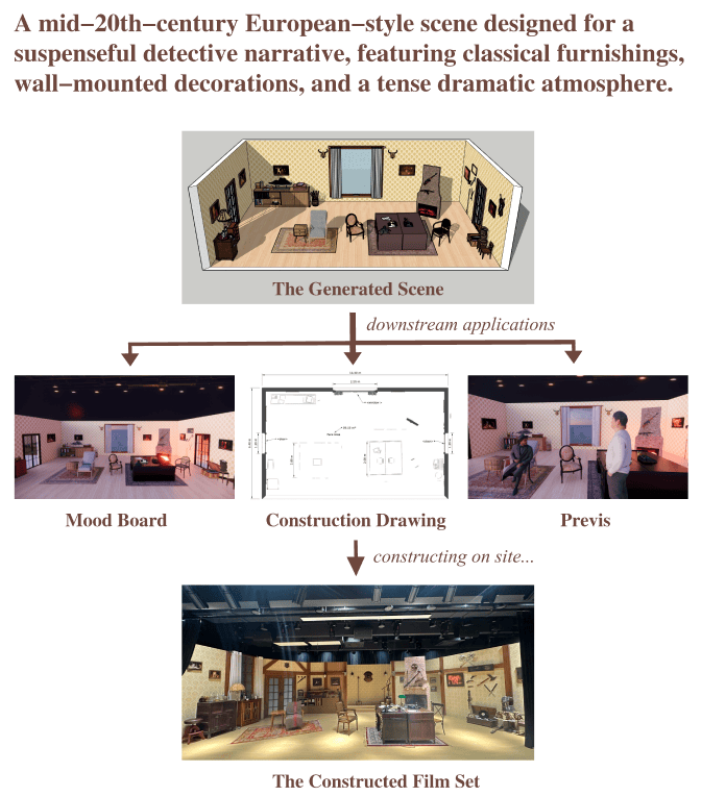}
\caption{Seamless integration of our system into film production workflows. Given a scene description, the system generates the corresponding scene efficiently. The director and the art department refine the generated scene, and create the construction drawings to construct the physical film set, while also utilizing it to support film previs and mood boards. Note: To accommodate lighting arrangements commonly used in film production, the roof is intentionally omitted during generation.}
  \label{fig:film_set}
\end{figure}

\noindent\textbf{Seamless Integration into Film Production Workflows.} 
Our system can be seamlessly integrated into real film production workflows, as illustrated in Figure~\ref{fig:film_set}. According to the director's requirements about film scene (\textit{A mid-20th-century European-style scene designed for a suspenseful detective narrative, featuring classical furnishings, wall-mounted decorations, and a tense dramatic atmosphere.}), the system can rapidly generate the corresponding scene in SketchUp. Then the director further refine and adjust the generated scene through interactive editing to meet artistic intentions. Once the scene is finalized, the art department further creates construction drawings, while also utilizing the generated scene to support film previs and mood boards. Finally, after approval by the director, the physical film set is constructed on site based on the construction drawings. Overall, our system fully integrates into traditional film production workflows and significantly improves the efficiency of film set design.

\section{Discussion}
\noindent\textbf{Limitation.} Although our system demonstrates the effectiveness of agent-based procedural generation for film set design, several limitations remain.  
First, our current system mainly focuses on scene generation and layout, without incorporating automated lighting. But lighting design is a critical component in film set construction, and our system does not consider integrating intelligent lighting strategies into the automated pipeline. 
Second, while our curated dataset SetDepot-Pro covers a wide range of film-specific assets, its scale can be further expanded. In particular, assets for specialized genres such as war and science fiction remain limited in the current version, and extending the dataset would enable broader applicability to diverse film productions.  
Third, as shown in Table 3 of main paper, the performance of our framework noticeably degrades when generating five or more rooms simultaneously. Although in real-world film set design it is uncommon to construct more than five rooms at once, addressing this scalability issue will improve the robustness and generality of our system.  

\noindent\textbf{Future Work.} In future work, we plan to improve the system along three directions: (1) incorporating automated lighting modules to better capture cinematic atmosphere, (2) continuously expanding SetDepot-Pro with assets for genres such as war and science fiction, and (3) optimizing agent collaboration mechanisms for multi-room reasoning to enhance performance in complex scenarios. These improvements will further increase the practicality of our system and align it more closely with professional film set design workflows.

\begin{figure*}[h]
  \centering
  \includegraphics[width=1\textwidth]{Figures/Adjacency.pdf}
  \caption{Prompts for the \textit{Adjacency} agent.}
  \label{fig:Adjacency}
\end{figure*}

\begin{figure*}[h]
  \centering
  \includegraphics[width=1\textwidth]{Figures/Material.pdf}
  \caption{Prompts for the \textit{Material} agent.}
  \label{fig:Material}
\end{figure*}

\begin{figure*}[h]
  \centering
  \includegraphics[width=1\textwidth]{Figures/Door_Window_part1.pdf}
  \caption{Prompts for the \textit{Door\_Window} agent (part1).}
  \label{fig:Door_Window_part1}
\end{figure*}

\begin{figure*}[h]
  \centering
  \includegraphics[width=1\textwidth]{Figures/Door_Window_part2.pdf}
  \caption{Prompts for the \textit{Door\_Window} agent (part2).}
  \label{fig:Door_Window_part2}
\end{figure*}

\begin{figure*}[h]
  \centering
  \includegraphics[width=1\textwidth]{Figures/Object_part1.pdf}
  \caption{Prompts for the \textit{Object} agent (part1).}
  \label{fig:Object_part1}
\end{figure*}

\begin{figure*}[h]
  \centering
  \includegraphics[width=1\textwidth]{Figures/Object_part2.pdf}
  \caption{Prompts for the \textit{Object} agent (part2).}
  \label{fig:Object_part2}
\end{figure*}


\end{document}